\def\year{2018}\relax
\documentclass[letterpaper]{article} 
\usepackage{aaai18}  
\usepackage{times}  
\usepackage{helvet}  
\usepackage{courier}  
\usepackage{url}  
\usepackage{graphicx}  

\usepackage{amsmath,amsbsy,amsfonts,amssymb,amsthm,dsfont,units, subcaption}
\usepackage{appendix}
\newcommand{\eg}{\emph{e.g.}}
\newcommand{\ie}{\emph{i.e.}}

\begin{document}
%
\title{Flow-GAN: Combining Maximum Likelihood and Adversarial Learning in Generative Models}
\author{Aditya Grover, Manik Dhar, Stefano Ermon\\
Computer Science Department\\
Stanford University\\
\texttt{\{adityag, dmanik, ermon\}@cs.stanford.edu}\\
}
\maketitle
\begin{abstract}
Adversarial learning of probabilistic models has recently emerged as a promising alternative to maximum likelihood. 
Implicit models such as generative adversarial networks (GAN) often generate better samples compared to explicit models trained by maximum likelihood. Yet, GANs sidestep the characterization of an explicit density which makes quantitative evaluations challenging. 
To bridge this gap, we propose Flow-GANs, a generative adversarial network for which we can perform \textit{exact} likelihood evaluation, thus supporting both adversarial and maximum likelihood training. 
When trained adversarially, Flow-GANs generate high-quality samples but attain extremely poor log-likelihood scores, inferior even to a mixture model memorizing the training data; the opposite is true when trained by maximum likelihood. 
Results on MNIST and CIFAR-10 demonstrate that hybrid training can attain high held-out likelihoods while retaining visual fidelity in the generated samples.
\end{abstract}

\section{Introduction}

Highly expressive parametric models have enjoyed great success in supervised learning, where learning objectives and evaluation metrics are typically well-specified and easy to compute. On the other hand, the learning objective for unsupervised settings is less clear. At a fundamental level, the idea is to learn a generative model that minimizes some notion of divergence with respect to the data distribution. Minimizing the Kullback-Liebler divergence between the data distribution and the model, for instance, is equivalent to performing maximum likelihood estimation (MLE) on the observed data.
Maximum likelihood estimators are asymptotically statistically efficient, and serve as natural objectives for learning \textit{prescribed generative models}~\cite{mohamed2016learning}. 

In contrast, an alternate principle that has recently attracted much attention is based on adversarial learning, where the objective is to generate data indistinguishable from the training data. Adversarially learned models such as generative adversarial networks (GAN;~\cite{goodfellow2014generative})  can sidestep specifying an explicit density for any data point and belong to the class of \textit{implicit generative models}~\cite{diggle1984monte}.

The lack of characterization of an explicit density in GANs is however problematic for two reasons. Several application areas of deep generative models rely on density estimates; for instance, count based exploration strategies based on density estimation using generative models have recently achieved state-of-the-art performance on challenging reinforcement learning environments~\cite{ostrovski2017count}. Secondly, it makes the quantitative evaluation of the generalization performance of such models
challenging. The typical evaluation criteria based on ad-hoc sample quality metrics~\cite{salimans2016improved,che2016mode} do not address this issue since it is possible to generate good samples by memorizing the training data, or missing important modes of the distribution, or both~\cite{theis2015note}. 
Alternatively, density estimates based on approximate inference techniques such as annealed importance sampling (AIS;~\cite{neal2001annealed,wu2016quantitative}) and non-parameteric methods such as kernel density estimation (KDE;~\cite{parzen1962estimation,goodfellow2014generative}) are computationally slow and 
crucially rely on assumptions of a Gaussian observation model for the likelihood that could lead to misleading estimates as we shall demonstrate in this paper.

To sidestep the above issues, we propose Flow-GANs,
a generative adversarial network with a normalizing flow generator. A Flow-GAN generator transforms a prior noise density into a model density through a sequence of invertible transformations. By using an invertible generator, Flow-GANs allow us to tractably evaluate \textit{exact} likelihoods using the change-of-variables formula and perform \textit{exact} posterior inference over the latent variables while still permitting efficient ancestral sampling, desirable properties of any probabilistic model that a typical GAN would not provide.

 Using a Flow-GAN, we perform a principled quantitative comparison of maximum likelihood and adversarial learning on benchmark datasets viz. MNIST and CIFAR-10. While adversarial learning outperforms MLE on sample quality metrics as expected based on strong evidence in prior work, the log-likelihood estimates of adversarial learning are orders of magnitude worse than those of MLE. The difference is so stark that a simple Gaussian mixture model baseline outperforms adversarially learned models on \textit{both} sample quality and held-out likelihoods. Our quantitative analysis reveals that the poor likelihoods of adversarial learning can be explained as a result of an ill-conditioned Jacobian matrix for the generator function suggesting a mode collapse, rather than overfitting to the training dataset.
 
To resolve the dichotomy of perceptually good-looking samples at the expense of held-out likelihoods in the case of adversarial learning (and vice versa in the case of MLE), we propose a hybrid objective that bridges implicit and prescribed learning by augmenting the adversarial training objective with an additional term corresponding to the log-likelihood of the observed data. While the hybrid objective achieves the intended effect of smoothly trading-off the two goals in the case of CIFAR-10, it has a regularizing effect on MNIST where it outperforms MLE and adversarial learning on both held-out likelihoods and sample quality metrics. 

Overall, this paper makes the following contributions:
\begin{enumerate}
\item We propose Flow-GANs, a generative adversarial network with an invertible generator that can perform efficient ancestral sampling and exact likelihood evaluation.
\item We propose a hybrid learning objective for Flow-GANs that attains good log-likelihoods and generates high-quality samples on MNIST and CIFAR-10 datasets.
\item We demonstrate the limitations of AIS and KDE for log-likelihood evaluation and ranking of implicit models.
\item We analyze the singular value distribution for the Jacobian of the generator function to explain the low log-likelihoods observed due to adversarial learning.
\end{enumerate}

\section{Preliminaries}

We begin with a review of maximum likelihood estimation and adversarial learning in the context of generative models. For ease of presentation, all distributions are w.r.t. any arbitrary  $\mathbf{x} \in \mathbb{R}^d$, unless otherwise specified. We use upper-case to denote probability distributions and assume they all admit absolutely continuous densities (denoted by the corresponding lower-case notation) on a reference measure $\mathrm{d}\mathbf{x}$. 

Consider the following setting for learning generative models. Given some data $X = \{ \mathbf{x}_i \in \mathbb{R}^d\}_{i=1}^m$ sampled i.i.d. from an unknown probability density $p_{\mathrm{data}}$, we are interested in learning a probability density $p_{\theta}$ where $\theta$ denotes the parameters of a  model. Given a parameteric family of models $\mathcal{M}$, the typical approach to learn $\theta \in \mathcal{M}$ is to minimize a notion of divergence between $P_{\mathrm{data}}$ and $P_{\theta}$. The choice of divergence and the optimization procedure dictate learning, leading to the following two objectives.

\subsection{Maximum likelihood estimation}
In maximum likelihood estimation (MLE), we minimize the Kullback-Liebler (KL) divergence between the data distribution and the model distribution. 
Formally, the learning objective can be expressed as:
\begin{align*}
\min_{\theta \in \mathcal{M}} KL(P_{\mathrm{data}}, P_{\theta}) &= \mathbb{E}_{\mathbf{x}\sim P_{\mathrm{data}}} \left[\log \frac{p_{\mathrm{data}} (\mathbf{x})}{p_\theta (\mathbf{x})}\right]
\end{align*}
Since $p_{data}$ is independent of $\theta$, the above optimization problem can be equivalently expressed as:
\begin{align}\label{eq:mle}
\max_{\theta \in \mathcal{M}} \mathbb{E}_{\mathbf{x}\sim P_{\mathrm{data}}} \left[\log p_\theta (\mathbf{x})\right]
\end{align}
Hence, evaluating the learning objective for MLE in Eq.~\eqref{eq:mle} requires the ability to evaluate the model density $p_\theta (\mathbf{x})$.  
Models that provide an explicit characterization of the likelihood function are referred to as prescribed generative models~\cite{mohamed2016learning}.

\subsection{Adversarial learning}
A generative model can be learned to optimize divergence notions beyond the KL divergence. A large family of divergences
can be conveniently expressed as:
\begin{align}\label{eq:div}
\max_{\phi \in \mathcal{F}} \;\; \mathbb{E}_{\mathbf{x} \sim P_{\theta}} \left[h_{\phi}(\mathbf{x})\right] -\mathbb{E}_{\mathbf{x} \sim P_{\mathrm{data}}} \left[h'_{\phi}(\mathbf{x})\right] 
\end{align}
where $\mathcal{F}$ denotes a set of parameters, $h_{\phi}$ and $h'_{\phi}$ are appropriate real-valued functions parameterized by $\phi$. Different choices of $\mathcal{F}$, $h_{\phi}$ and $h'_{\phi}$ can lead to a variety of $f$-divergences such as Jenson-Shannon divergence and integral probability metrics such as the Wasserstein distance. For instance, the GAN objective proposed by \citeauthor{goodfellow2014generative}~\shortcite{goodfellow2014generative} can also be cast in the form of Eq.~\eqref{eq:div} below:
\begin{align}\label{eq:jsd}
\max_{\phi \in \mathcal{F}} \;\; \mathbb{E}_{\mathbf{x} \sim P_{\theta}} \left[\log \left(1-D_{\phi}(\mathbf{x})\right)\right] +\mathbb{E}_{\mathbf{x} \sim P_{\mathrm{data}}} \left[D_{\phi}(\mathbf{x})\right] 
\end{align}
where $\phi$ denotes the parameters of a neural network function $D_\phi$.
We refer the reader to~\cite{nowozin2016f,mescheder2017numerics} for further details on other possible choices of divergences. Importantly, a Monte Carlo estimate of the objective in Eq.~\eqref{eq:div} requires only samples from the model.
Hence, any model that allows tractable sampling can be used to evaluate the following minimax objective:
\begin{align}\label{eq:minimax}
\min_{\theta\in \mathcal{M}} \max_{\phi \in \mathcal{F}} \;\; & \mathbb{E}_{\mathbf{x} \sim P_{\theta}} \left[h_{\phi}(\mathbf{x})\right] -\mathbb{E}_{\mathbf{x} \sim P_{\mathrm{data}}} \left[h'_{\phi}(\mathbf{x})\right].
\end{align}

As a result, even differentiable \textit{implicit models} which do not provide a characterization of the model likelihood\footnote{\label{foot}This could be either due to computational intractability in evaluating likelihoods or because the likelihood is ill-defined.} but allow tractable sampling can be learned adversarially by optimizing minimax objectives of the form given in Eq.~\eqref{eq:minimax}.

\subsection{Adversarial learning of latent variable models}
From a statistical perspective, maximum likelihood estimators are statistically efficient asymptotically (under some conditions) and hence minimizing the KL divergence is a natural objective for many prescribed models~\cite{huber1967behavior}. However, not all models allow for a well-defined, tractable, and easy-to-optimize likelihood.

For example, exact likelihood evaluation and sampling are tractable in directed, fully observed models such as Bayesian networks and
autoregressive models~\cite{larochelle2011neural,oord2016pixel}. Hence, they are usually trained by maximum likelihood. Undirected models,
on the other hand, provide only unnormalized likelihoods and are sampled from using expensive Markov chains. Hence, they are usually learned by approximating the likelihood using methods such as contrastive divergence~\cite{carreira2005contrastive} and pseudolikelihood~\cite{besag1977efficiency}. The likelihood is generally intractable to compute in latent variable models (even directed ones) as it requires marginalization. 
These models are typically learned by optimizing a stochastic lower bound to the log-likelihood using variational Bayes approaches~\cite{kingma-iclr2014}. 

Directed latent variable models allow for efficient ancestral sampling and hence these models can also be trained using other divergences, \eg, adversarially~\cite{mescheder2017adversarial,mao2016least,song2017nice}. A popular class of latent variable models learned adversarially consist of generative adversarial networks 
(GAN;~\cite{goodfellow2014generative}). GANs comprise of a pair of generator and discriminator networks.
The generator $G_\theta: \mathbb{R}^k \rightarrow \mathbb{R}^d$ is a deterministic function differentiable with respect to the parameters $\theta$. The function takes as input a source of randomness $\mathbf{z} \in \mathbb{R}^k$ sampled from a tractable prior density $p(\mathbf{z})$ and transforms it to a sample $G_\theta(\mathbf{z})$ through a forward pass. Evaluating likelihoods assigned by a GAN is challenging because the model density $p_\theta$ is specified only implicitly using the prior density $p(\mathbf{z})$ and the generator function $G_\theta$. In fact, the likelihood for any data point is ill-defined (with respect to the Lesbegue measure over $\mathbb{R}^n$) if the prior distribution over $\mathbf{z}$ is defined over a support smaller than the support of the data distribution. 

GANs are typically learned adversarially with the help of a discriminator network. The discriminator $D_\phi: \mathbb{R}^d \rightarrow \mathbb{R}$ is another real-valued function that is differentiable with respect to a set of parameters $\phi$. Given the discriminator function, we can express the functions $h$ and $h'$ in Eq.~\eqref{eq:minimax} as compositions of  $D_\phi$ with divergence-specific functions.  For instance, the Wasserstein GAN (WGAN;~\cite{arjovsky2017wasserstein}) optimizes the following objective:

\begin{align}\label{eq:wgan}
\min_\theta \max_{\phi \in \mathcal{F}} \mathbb{E}_{\mathbf{x} \sim P_{\mathrm{data}}} \left[D_\phi(\mathbf{x})\right] - \mathbb{E}_{\mathbf{z} \sim P_{\mathbf{z}}} \left[ D_\phi(G_\theta(\mathbf{z})) \right]
\end{align}
where $\mathcal{F}$ is defined such that $D_\phi$ is 1-Lipschitz. Empirically, GANs generate excellent samples of natural images~\cite{radford2015unsupervised}, audio signals~\cite{pascual2017segan}, and of behaviors in imitation learning~\cite{ho2016generative,li2017inferring}.

\section{Flow Generative Adversarial Networks}

As discussed above, generative adversarial networks can tractably generate high-quality samples but have intractable or ill-defined likelihoods. Monte Carlo techniques such as AIS and non-parameteric density estimation methods such as KDE get around this by assuming a Gaussian observation model $p_\theta(\mathbf{x} \vert \mathbf{z})$ for the generator.\footnote{The true observation model for a GAN is a Dirac delta distribution, \ie, $p_\theta(\mathbf{x} \vert \mathbf{z})$ is infinite when $\mathbf{x} = G_\theta(\mathbf{z})$ and zero otherwise.} This assumption alone is not sufficient for quantitative evaluation since the marginal likelihood of the observed data, $p_\theta(\mathbf{x}) = \int p_\theta(\mathbf{x}, \mathbf{z}) \mathrm{d}\mathbf{z}$ in this case would be intractable as it requires integrating over all the latent factors of variation. This would then require approximate inference (\eg, Monte Carlo or variational methods) which in itself is a computational challenge for high-dimensional distributions. To circumvent these issues, we propose flow generative adversarial networks (Flow-GAN). 

A Flow-GAN consists of a pair of generator-discriminator networks with the generator specified as a normalizing flow model~\cite{dinh2014nice}. A normalizing flow model specifies a parametric transformation from a prior density $p(\mathbf{z}): \mathbb{R}^d \rightarrow \mathbb{R}_0^+$ to another density over the same space, $p_\theta(\mathbf{x}): \mathbb{R}^d \rightarrow \mathbb{R}_0^+$ where $\mathbb{R}_0^+$ is the set of non-negative reals. The generator transformation $G_\theta: \mathbb{R}^d \rightarrow \mathbb{R}^d$ is invertible, such that there exists an inverse function $f_\theta=G_\theta^{-1}$. Using the change-of-variables formula and letting $\mathbf{z}=f_\theta(\mathbf{x})$, we have: 
\begin{align}\label{eq:change_of_var}
p_\theta (\mathbf{x}) = p(\mathbf{z}) \left \vert \mathrm{det} \frac{\partial f_\theta(\mathbf{x})}{\partial \mathbf{x}} \right \vert
\end{align} 
where $\frac{\partial f_\theta(\mathbf{x})}{\partial \mathbf{x}}$ denotes the Jacobian of $f_\theta$ at $\mathbf{x}$. The above formula can be applied recursively over compositions of many invertible transformations to produce a complex final density. Hence, we can evaluate and optimize for the log-likelihood assigned by the model to a data point as long as the prior density is tractable and the determinant of the Jacobian of $f_\theta$ evaluated at $\mathbf{x}$ can be efficiently computed. 

Evaluating the likelihood assigned by a Flow-GAN model in Eq.~\eqref{eq:change_of_var} requires overcoming two major challenges. First, requiring the generator function $G_\theta$ to be reversible imposes a constraint on the dimensionality of the latent variable $\mathbf{z}$ to match that of the data $\mathbf{x}$. Thereafter, we require the transformations between the various layers of the generator to be invertible such that their overall composition results in an invertible $G_\theta$. Secondly, the Jacobian of high-dimensional distributions can however be computationally expensive to compute. If the transformations are designed such that the Jacobian is an upper or lower triangular matrix, then the determinant can be easily evaluated as the product of its diagonal entries. We consider two such family of transformations.

\begin{enumerate}
\item \textit{Volume preserving transformations}. Here, the Jacobian of the transformations have a unit determinant. For example, the NICE model consists of several layers performing a location transformation~\cite{dinh2014nice}. The top layer is a diagonal scaling matrix with non-zero log determinant.
\item \textit{Non-volume preserving transformations}. The determinant of the Jacobian of the transformations is not necessarily unity. For example, in Real-NVP, layers performs both location and scale transformations~\cite{dinh2016density}.
\end{enumerate}

For brevity, we direct the reader to \citeauthor{dinh2014nice}~\shortcite{dinh2014nice} and \citeauthor{dinh2016density}~\shortcite{dinh2016density} for the specifications of NICE and Real-NVP respectively. Crucially, both volume preserving and non-volume preserving transformations  are invertible such that the determinant of the Jacobian can be computed tractably. 
\begin{figure*}
\centering
\begin{subfigure}[b]{0.33\textwidth}
\centering
\includegraphics[width=0.7\textwidth]{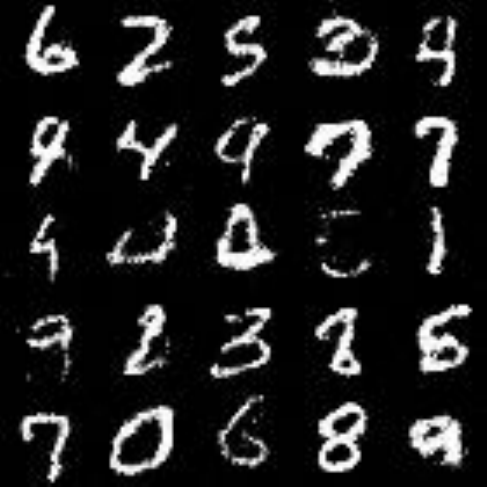}
\caption*{}
\end{subfigure}
\begin{subfigure}[b]{0.33\textwidth}
\centering
\includegraphics[width=0.7\columnwidth]{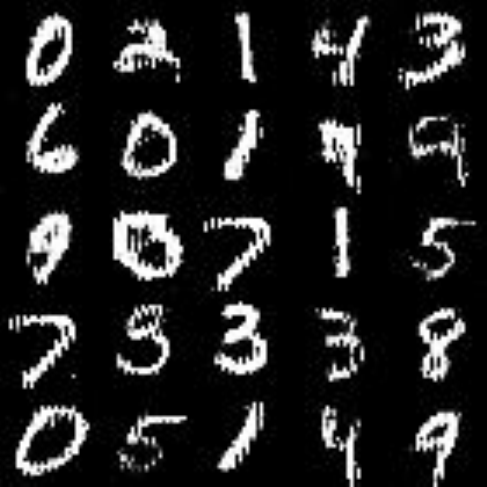}
\caption*{}
\end{subfigure}
\begin{subfigure}[b]{0.33\textwidth}
\centering
\includegraphics[width=0.7\columnwidth]{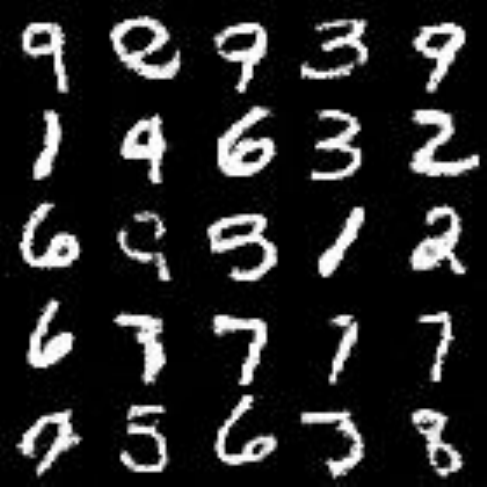}
\caption*{}
\end{subfigure}

\begin{subfigure}[b]{0.33\textwidth}
\centering
\includegraphics[width=0.7\columnwidth]{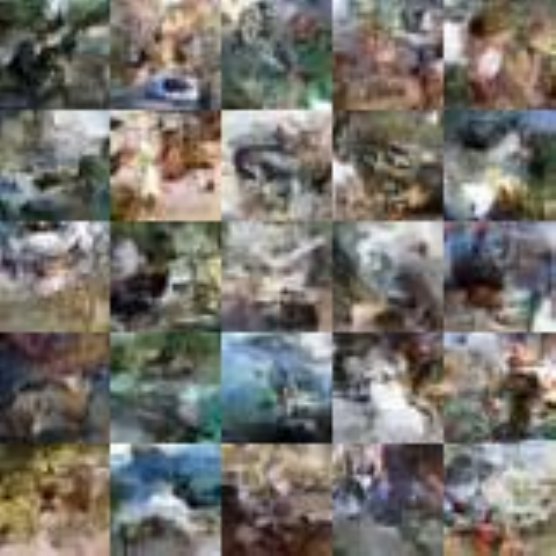}
\caption{MLE}\label{fig:samples_mle}
\end{subfigure}
\begin{subfigure}[b]{0.33\textwidth}
\centering
\includegraphics[width=0.7\columnwidth]{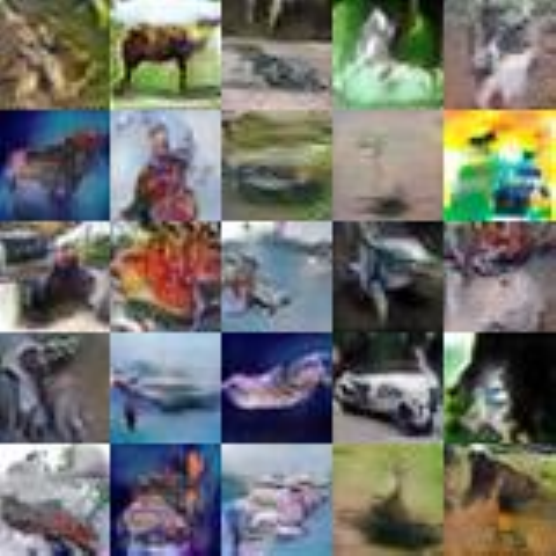}
\caption{ADV}\label{fig:samples_adv}
\end{subfigure}
\begin{subfigure}[b]{0.33\textwidth}
\centering
\includegraphics[width=0.7\columnwidth]{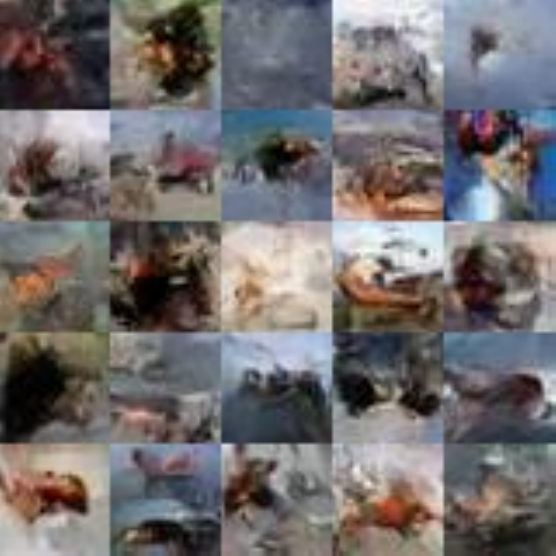}
\caption{Hybrid}\label{fig:samples_hybrid}
\end{subfigure}
\caption{Samples generated by Flow-GAN models with different objectives for MNIST (\textbf{top}) and CIFAR-10 (\textbf{bottom}).}\label{fig:mnist_sample}
\end{figure*}

\subsection{Learning objectives}
In a Flow-GAN, the likelihood is well-defined and computationally tractable for exact evaluation of even expressive volume preserving and non-volume preserving transformations. Hence, a Flow-GAN can be trained via maximum likelihood estimation using Eq.~\eqref{eq:mle} in which case the discriminator is redundant. Additionally, we can perform ancestral sampling just like a regular GAN whereby we sample a random vector $\mathbf{z}\sim P_z$ and transform it to a model generated sample via $G_\theta=f_\theta^{-1}$.  This makes it possible to learn a Flow-GAN using an adversarial learning objective (for example, the WGAN objective in Eq.~\eqref{eq:wgan}).

A natural question to ask is why should one use adversarial learning given that MLE is  statistically efficient asymptotically (under some conditions). 
Besides difficulties that could arise due to optimization (in both MLE and adversarial learning), the optimality of MLE holds only when there is no model misspecification for the generator \ie, the true data distribution $P_{\mathrm{data}}$ is a member of the parametric family of distributions under consideration~\cite{white1982maximum}.
This is generally not the case for high-dimensional distributions, and hence the choice of the learning objective becomes largely an empirical question. Unlike other models, a Flow-GAN allows both maximum likelihood and adversarial learning, and hence we can investigate this question experimentally.

\subsection{Evaluation metrics and experimental setup}
Our criteria for evaluation is based on held-out log-likelihoods and sample quality metrics. We focus on natural images since they allow visual inspection as well as quantification using recently proposed metrics. 
A ``good'' generative model should generalize to images outside the training data and assign high log-likelihoods to held-out data.
The Inception and MODE scores are standard quantitative measures of the quality of 
generated samples of natural images for labelled datasets~\cite{salimans2016improved,che2016mode}. The Inception scores are computed as:
\begin{align*}
\exp\left (\mathbb{E}_{\mathbf{x} \in P_\theta} [KL(p(y \vert \mathbf{x}) \Vert p(y)]  \right)
\end{align*}
where $\mathbf{x}$ is a sample generated by the model, $p(y\vert \mathbf{x})$ is the softmax probability for the labels $y$ assigned by a pretrained classifier for $\mathbf{x}$, and $p(y)$ is the overall distribution of labels in the generated samples (as predicted by the pretrained classifier). The intuition is that the conditional distribution $p(y\vert \mathbf{x})$ should have low entropy for good looking images while the marginal distribution $p(y)$ has high entropy to ensure sample diversity. Hence, a generative model can perform well on this metric if the KL divergence  between the two distributions (and consequently, the Inception score for the generated samples) is large. 
The MODE score given below modifies the Inception score to take into account the distribution of labels in the training data, $p^\ast(y)$:
\begin{align*}
\exp\left (\mathbb{E}_{\mathbf{x} \in P_\theta} [KL(p(y \vert \mathbf{x}) \Vert p^\ast(y)] - KL(p^\ast(y) \Vert p(y)) \right).
\end{align*}

We compare learning of Flow-GANs using MLE and adversarial learning (ADV) for the MNIST dataset of handwritten digits~\cite{lecun2010mnist} and the CIFAR-10 dataset of natural images~\cite{krizhevsky2009learning}. 
The normalizing flow generator architectures are chosen to be NICE~\cite{dinh2014nice} and Real-NVP~\cite{dinh2016density} for MNIST and CIFAR-10 respectively. We fix the Wasserstein distance as the choice of the divergence being optimized by ADV (see Eq.~\eqref{eq:wgan}) with the Lipschitz constraint over the critic imposed by penalizing the norm of the gradient with respect to the input~\cite{arjovsky2017wasserstein,gulrajani2017improved}. The discriminator is based on the DCGAN architecture~\cite{radford2015unsupervised}. The above choices are among the current state-of-the-art in maximum likelihood estimation and adversarial learning and greatly stabilize GAN training.
Further experimental setup details are provided in Appendix~\ref{app:exp_setup}. The code for reproducing the results is available at \texttt{https://github.com/ermongroup/flow-gan}.

\begin{figure}[th]
\centering
\begin{subfigure}[b]{0.48\columnwidth}
\centering
\includegraphics[width=\columnwidth]{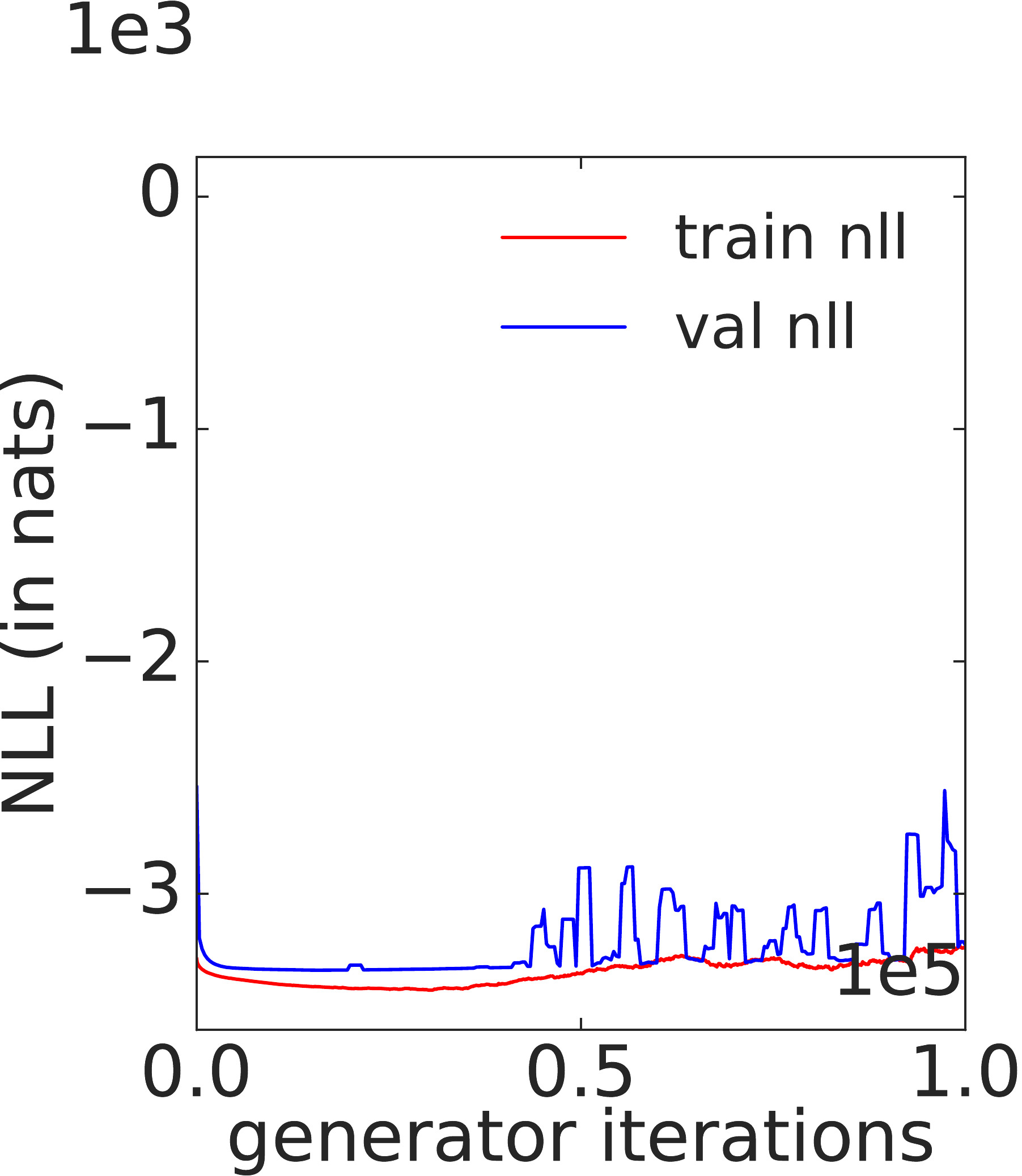}
\caption*{}
\end{subfigure}
\begin{subfigure}[b]{0.48\columnwidth}
\centering
\includegraphics[width=\columnwidth]{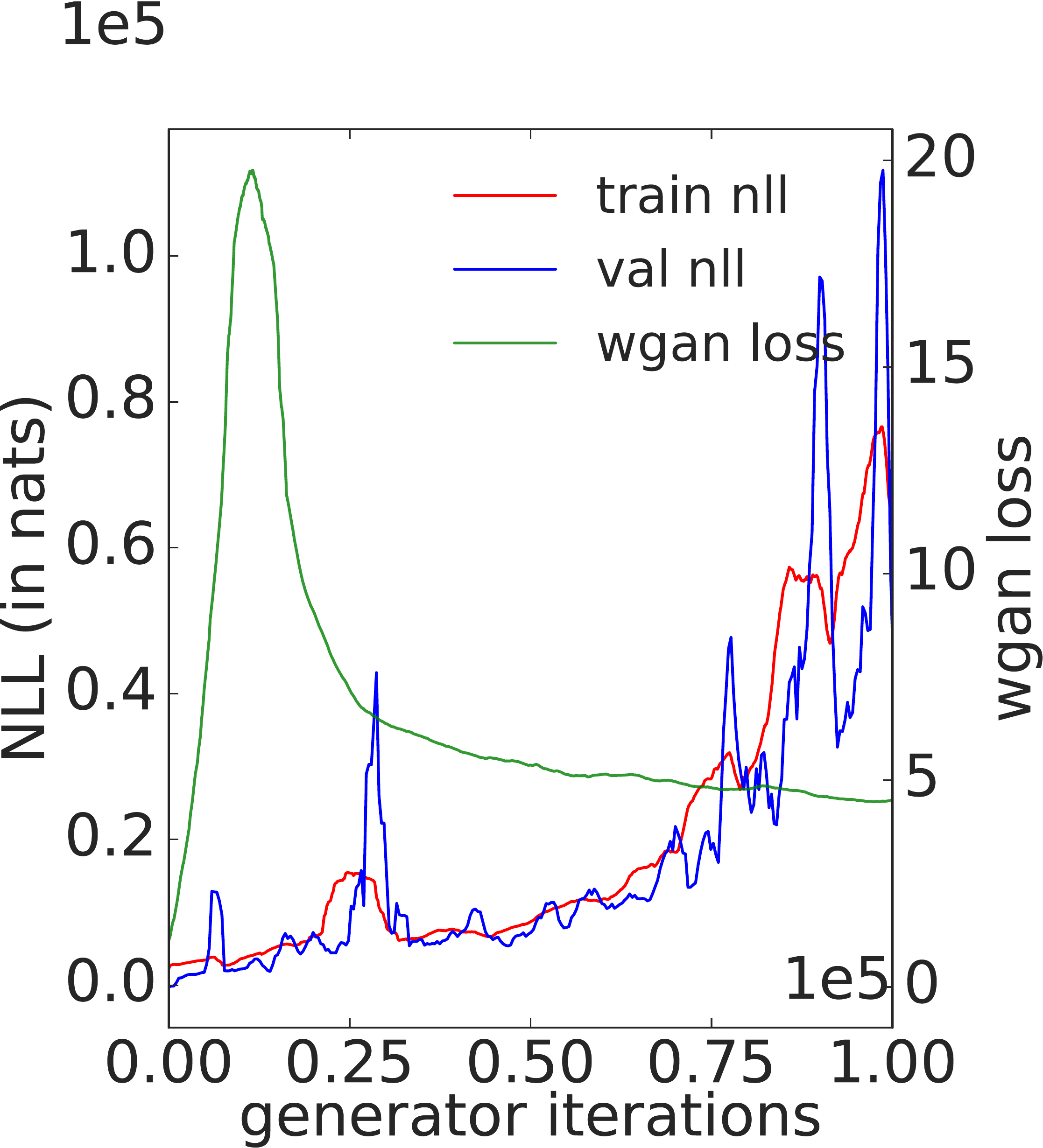}
\caption*{}
\end{subfigure}
\begin{subfigure}[b]{0.495\columnwidth}
\centering
\includegraphics[width=\columnwidth]{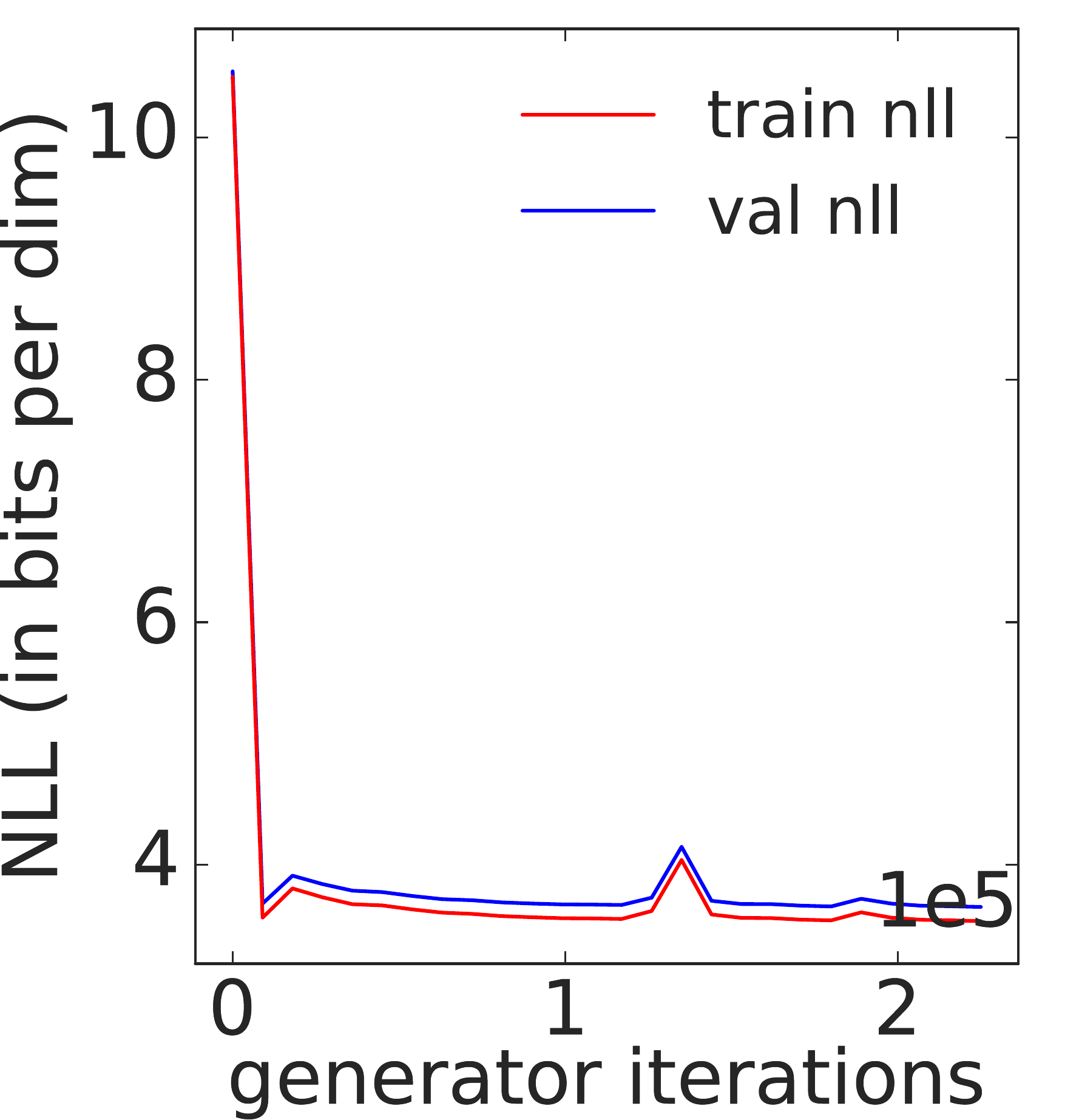}
\caption{MLE}\label{fig:nll_mle}
\end{subfigure}
\begin{subfigure}[b]{0.495\columnwidth}
\centering
\includegraphics[width=\columnwidth]{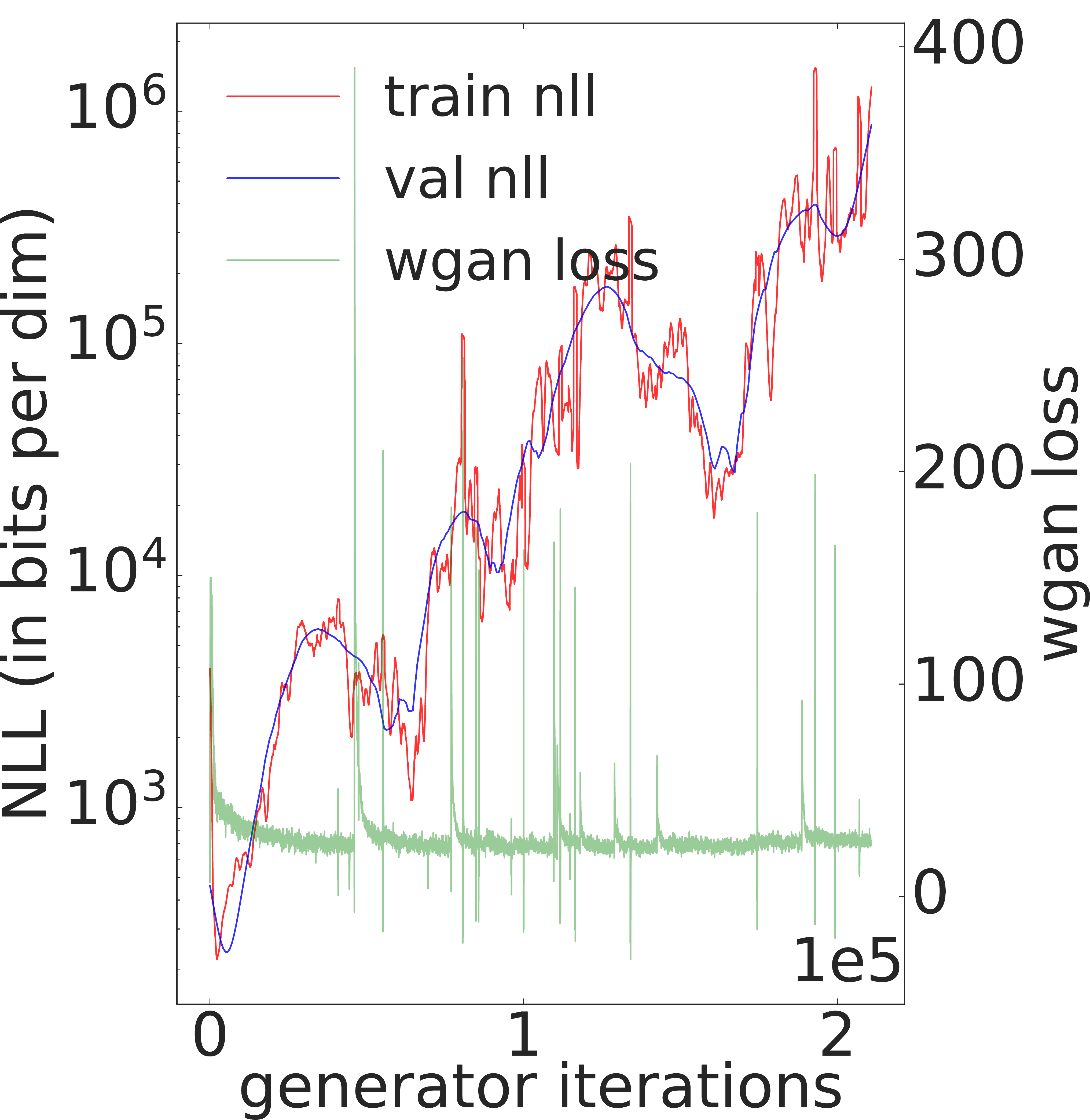}
\caption{ADV}\label{fig:nll_adv}
\end{subfigure}
\caption{Learning curves for negative log-likelihood (NLL) evaluation on MNIST (\textbf{top}, in nats) and CIFAR (\textbf{bottom}, in bits/dim). Lower NLLs are better.
}\label{fig:mnist_nll}
\end{figure}

\subsection{Evaluation results}
\paragraph{Log-likelihood.}

The log-likelihood learning curves for  Flow-GAN models learned using MLE and ADV are shown in Figure~\ref{fig:nll_mle} and Figure~\ref{fig:nll_adv} respectively. Following convention, we report the negative log-likelihoods (NLL) in nats for MNIST and bits/dimension for CIFAR-10. 

\emph{\textbf{MLE.}} In Figure~\ref{fig:nll_mle}, we see that normalizing flow models attain low validation NLLs (blue curves) after few gradient updates as expected because it is explicitly optimizing for the MLE objective in Eq.~\eqref{eq:mle}. Continued training however could lead to overfitting as the train NLLs (red curves) begin to diverge from the validation NLLs.

\emph{\textbf{ADV.}} Surprisingly, ADV models show a consistent \textit{increase} in validation NLLs as training progresses  as shown in Figure~\ref{fig:nll_adv} (for  CIFAR-10, the estimates are reported on a log scale!). Based on the learning curves, we can disregard overfitting as an explanation since the increase in NLLs is observed even on the training data. The training and validation NLLs closely track each other suggesting that ADV models are not simply memorizing the training data. 

Comparing the left vs. right panels in Figure~\ref{fig:mnist_nll}, we see that the log-likelihoods attained by ADV are orders of magnitude worse than those attained by MLE after sufficient training. 
Finally, we note that the WGAN loss (green curves) does not correlate well with NLL estimates. While the WGAN loss stabilizes after few iterations of training, the NLLs continue to increase.
This observation is in contrast to prior work showing the loss to be strongly correlated with sample quality metrics~\cite{arjovsky2017wasserstein}. 

\paragraph{Sample quality.}
Samples generated from MLE and ADV-based models with the best MODE/Inception are shown in Figure~\ref{fig:samples_mle} and Figure~\ref{fig:samples_adv} respectively.
ADV models significantly outperform MLE with respect to the final MODE/Inception scores achieved. Visual inspection of samples confirms the observations made on the based of the sample quality metrics.
Curves monitoring the sample quality metrics at every training iteration are given in Appendix~\ref{app:sample_quality}.

\subsection{Gaussian mixture models }
\begin{figure}[t]
\centering
\includegraphics[width=0.8\columnwidth]{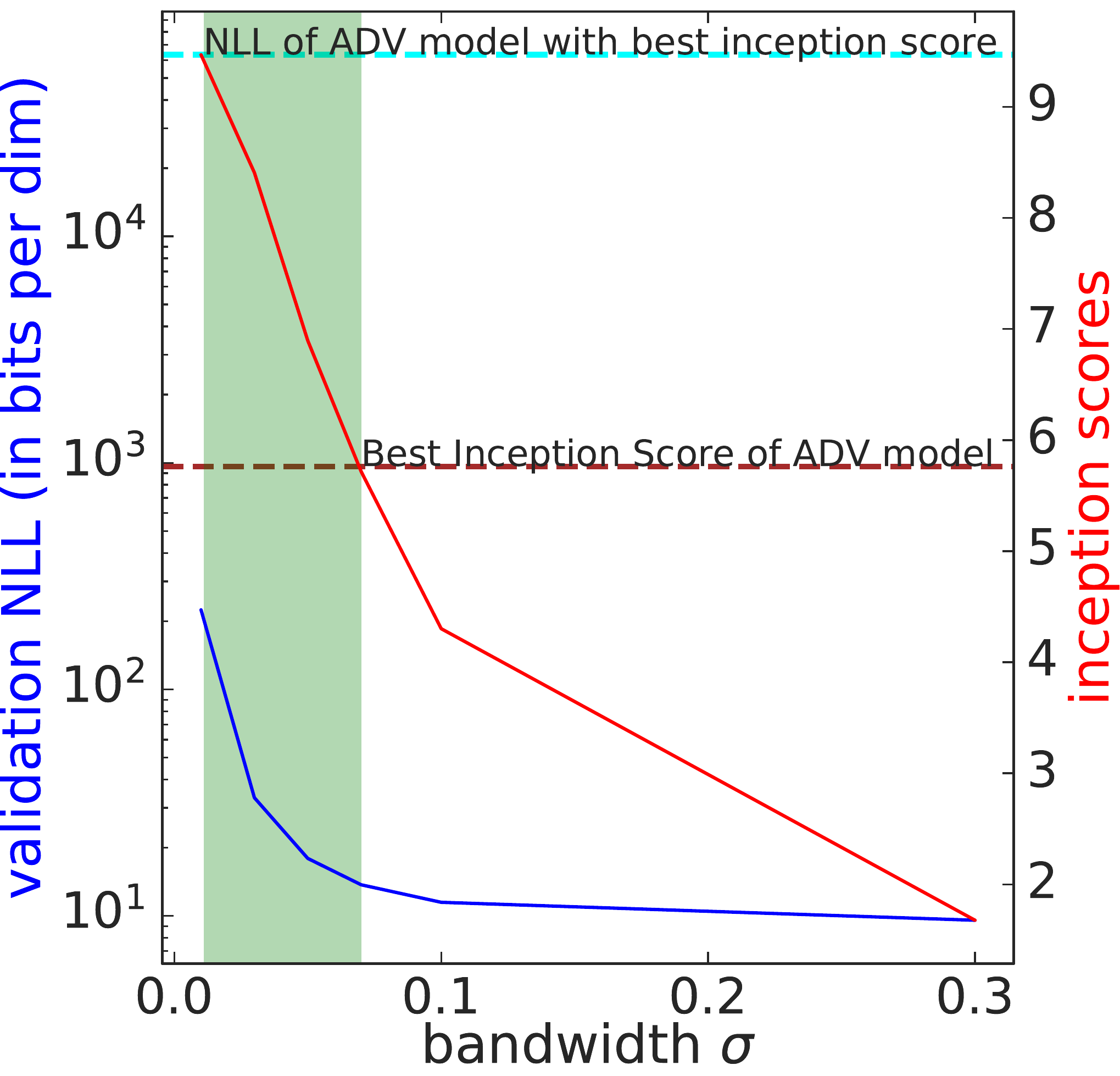}
\caption{Gaussian Mixture Models outperform adversarially learned models on both held-out log-likelihoods and sampling metrics on CIFAR-10 (\textbf{green shaded region}).}\label{fig:gmm}
\end{figure}

The above experiments suggest that ADV can produce excellent samples but assigns low likelihoods to the observed data. However, a direct comparison of ADV with the log-likelihoods of MLE is unfair since the latter is explicitly optimizing for the desired objective. To highlight that generating good samples at the expense of low likelihoods is \emph{not} a challenging goal, we propose a simple baseline. We compare the adversarially learned Flow-GAN models that achieves the highest MODE/Inception score with a Gaussian Mixture Model consisting of $m$ isotropic Gaussians with equal weights centered at each of the $m$ training points as the baseline Gaussian Mixture Model (GMM). The bandwidth hyperparameter, $\sigma$, is the same for each of the mixture components and optimized for the lowest validation NLL by doing a line search in $(0, 1]$. We show results for CIFAR-10 in Figure~\ref{fig:gmm}. Our observations below hold for MNIST as well; results deferred to Appendix~\ref{app:gmm}.

\
We overload the $y$-axis in Figure~\ref{fig:gmm} to report both NLLs and sample quality metrics. The horizontal maroon and cyan dashed lines denote the best attainable MODE/Inception scores and corresponding validation NLLs respectively attained by the adversarially learned Flow-GAN model. The GMM can clearly attain better sample quality metrics since it is explicitly overfitting to the training data for low values of the bandwidth parameter (any $\sigma$ for which the red curve is above the maroon dashed line). 
Surprisingly, the simple GMM also outperforms the adversarially learned model with respect to NLLs attained for several values of the bandwidth parameter (any $\sigma$ for which the blue curve is below the cyan dashed line). Bandwidth parameters for which GMM models outperform the adversarially learned model on both log-likelihoods and sample quality metrics are highlighted using the green shaded area. We show samples from the GMM in the appendix. Hence, \emph{a trivial baseline that is memorizing the training data can generate high quality samples and better held-out log-likelihoods}, suggesting that the log-likelihoods attained by adversarial training are very poor.

\section{Hybrid learning of Flow-GANs}
In the previous section, we observed that adversarially learning Flow-GANs models attain poor held-out log-likelihoods. This makes it challenging to use such models for applications requiring density estimation. On the other hand, Flow-GANs learned using MLE are ``mode covering" but do not generate high quality samples. With a Flow-GAN, it is possible to trade-off the two goals by combining the learning objectives corresponding to both these inductive principles. Without loss of generality, let $V(G_\theta, D_\phi)$ denote the minimax objective of any GAN model (such as WGAN). The hybrid objective of a Flow-GAN can be expressed as:
\begin{align}\label{eq:hybrid}
\min_\theta \max_\phi V(G_\theta,D_\phi) - \lambda \mathbb{E}_{\mathbf{x}\sim P_{\mathrm{data}}}\left[ \log p_\theta (\mathbf{x})\right] 
\end{align}
where $\lambda \geq 0$ is a hyperparameter for the algorithm. By varying $\lambda$, we can interpolate between plain adversarial training ($\lambda = 0$) and MLE (very high $\lambda$).

We summarize the results from MLE, ADV, and Hybrid for log-likelihood and sample quality evaluation in Table~\ref{tab:mnist} and Table~\ref{tab:cifar} for MNIST and CIFAR-10 respectively. The tables report the test log-likelihoods corresponding to the best validated MLE and ADV models and the highest MODE/Inception scores observed during training. The samples generated by models with the best MODE/Inception scores for each objective are shown in Figure~\ref{fig:samples_hybrid}. 

While the results on CIFAR-10 are along expected lines, the hybrid objective interestingly outperforms MLE and ADV on both test log-likelihoods and sample quality metrics in the case of MNIST. One potential explanation for this is that the ADV objective can regularize MLE to generalize to the test set and in turn, the MLE objective can stabilize the optimization of the ADV objective. Hence, the hybrid objective in Eq.~\eqref{eq:hybrid} can smoothly balance the two objectives using the tunable hyperparameter $\lambda$, and in some cases such as MNIST, the performance on both tasks could improve as a result of the hybrid objective.

\section{Interpreting the results}
Our findings are in contrast with prior work 
which report much better log-likelihoods for adversarially learned models with a standard generator architecture based on annealed importance sampling (AIS; \cite{wu2016quantitative}) and kernel density estimation (KDE; ~\cite{goodfellow2014generative}). These methods rely on approximate inference techniques for log-likelihood evaluation and make assumptions about a Gaussian observation model which does not hold for GANs.
Since Flow-GANs allow us to compute \emph{exact} log-likelihoods, we can evaluate the quality of approximation made by AIS and KDE for density estimation of invertible generators.
For a detailed description of the methods, we refer the reader to prior work~\cite{neal2001annealed,parzen1962estimation}.
 
\begin{table}[t]
\centering
\caption{Best MODE scores and test negative log-likelihood estimates for Flow-GAN models on MNIST.}
\label{tab:mnist}
\begin{tabular}{l|c|c}
Objective         & MODE Score & Test NLL  (in nats)        \\\hline 
MLE     & $7.42$  & $-3334.56$    \\
ADV     & $9.24$ & $-1604.09$  \\
Hybrid ($\lambda=0.1$) & $\mathbf{9.37}$ &  $\mathbf{-3342.95}$  
\end{tabular}
\end{table}

\begin{table}[t]
\centering
\caption{Best Inception scores and test negative log-likelihood estimates for  Flow-GAN models on CIFAR-10.}
\label{tab:cifar}
\begin{tabular}{l|c|c}
Objective         & Inception Score& Test NLL (in bits/dim)  \\\hline 
MLE   
  & $2.92$  &  $\mathbf{3.54}$   \\
ADV   
  & $\mathbf{5.76}$  &  $8.53$ \\
Hybrid ($\lambda=1$) & $3.90$  & $4.21$  
\end{tabular}
\end{table}

\begin{table}[t]
\centering
\caption{Comparison of inference techniques for negative log-likelihood estimation of Flow-GAN models on MNIST.}
\label{tab:approx}
\begin{tabular}{l|c|c|c}
Objective         & Flow-GAN NLL & AIS &  KDE       \\\hline 
MLE     & -3287.69  & -2584.40 & -167.10   \\
ADV     & 26350.30 & -2916.10  & -3.03\\
Hybrid & -3121.53 &  -2703.03  & -205.69
\end{tabular}
\end{table}

We consider the MNIST dataset where these methods have been previously applied to by~\citeauthor{wu2016quantitative}~\shortcite{wu2016quantitative} and~\citeauthor{goodfellow2014generative}~\shortcite{goodfellow2014generative} respectively. Since both AIS and KDE inherently rely on the samples generated, we evaluate these methods for the MLE, ADV, and Hybrid Flow-GAN model checkpoints corresponding to the best MODE scores observed during training. 
In Table~\ref{tab:approx}, we observe that both AIS and KDE produce estimates of log-likelihood that are far from the ground truth, accessible through the exact Flow-GAN log-likelihoods.
Even worse, the ranking of log-likelihood estimates for AIS (ADV$>$Hybrid$>$MLE) and KDE (Hybrid$>$MLE$>$ADV) do not obey the  relative rankings of the Flow-GAN estimates (MLE$>$Hybrid$>$ADV).  

\subsection{Explaining log-likelihood trends}
\begin{figure}[ht]
\centering
\includegraphics[width=0.75\columnwidth]{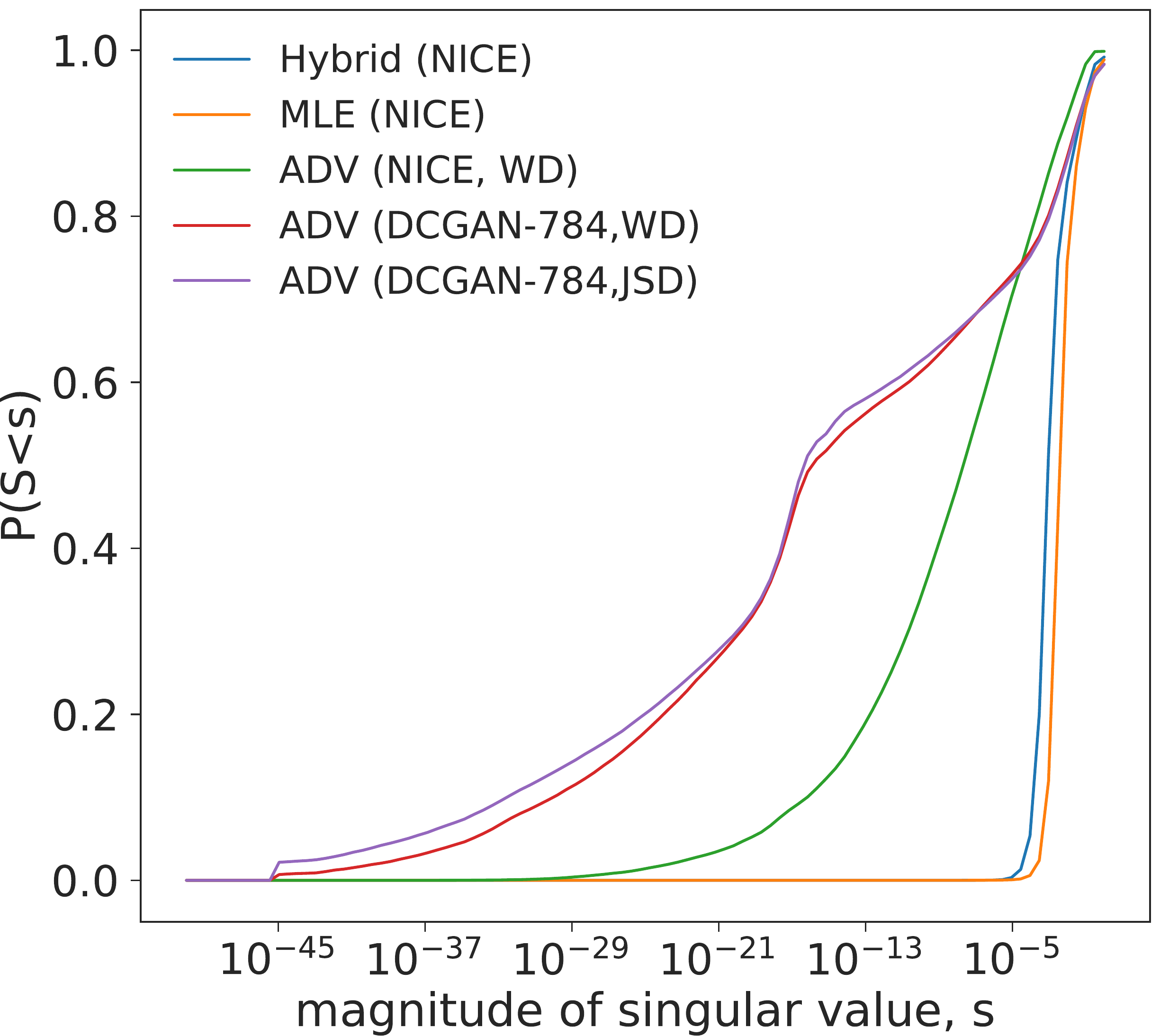}
\caption{CDF of the singular values magnitudes for the Jacobian of the generator functions trained on MNIST.}\label{fig:jac}
\end{figure}

In order to explain the variation in log-likelihoods attained by various Flow-GAN learning objectives, we investigate the distribution of the magnitudes of singular values for the Jacobian matrix of several generator functions, $G_\theta$ for MNIST in Figure~\ref{fig:jac} evaluated at 64 noise vectors $\mathbf{z}$ randomly sampled from the prior density $p(\mathbf{z})$. The $x$-axis of the figure shows the singular value magnitudes on a log scale and for each singular value $s$, we show the corresponding cumulative distribution function value on the $y$-axis which signifies the fraction of singular values less than $s$.
The results on CIFAR-10 in Appendix~\ref{app:jac} show a similar trend.

The Jacobian is a good first-order approximation of the generator function locally. In Figure~\ref{fig:jac}, we observe that the singular value distribution for the Jacobian of an invertible generator learned using MLE (orange curves) is concentrated in a narrow range, and hence the Jacobian matrix is well-conditioned and easy to invert. In the case of invertible generators learned using ADV with Wasserstein distance (green curves) however, the spread of singular values is very wide, and hence the Jacobian matrix is ill-conditioned. 

The average log determinant of the Jacobian matrices for MLE, ADV, and Hybrid models are $-4170.34, -15588.34$, and $-5184.40$ respectively which translates to the trend ADV$<$Hybrid$<$MLE. This indicates that the ADV models are trying to squish a sphere of unit volume centered at a latent vector $\mathbf{z}$ to a very small volume in the observed space $\mathbf{x}$. 
Tiny perturbations of training as well as held-out datapoints can
 hence manifest as poor log-likelihoods. In spite of not being limited in the representational capacity to cover the entire space of the data distribution (the dimensions of $\mathbf{z}$ (\ie, $k$) and $\mathbf{x}$ (\ie, $d$) match for invertible generators), ADV prefers to learn a distribution over a smaller support. 

The Hybrid learning objective (blue curves), however, is able to correct for this behavior, and the distribution of singular value magnitudes matches closely to that of MLE.  
We also considered variations involving the standard DCGAN architectures with $k=d$ minimizing the Wasserstein distance (red curves) and Jenson-Shannon divergence (purple curves). The relative shift in distribution of singular value magnitudes to lower values is apparent even in these cases.

\section{Discussion}
Any model which allows for efficient likelihood evaluation and sampling can be trained using maximum likelihood and adversarial learning.
This line of reasoning has been explored to some extent in prior work that combine the objectives of prescribed latent variable models such as VAEs (maximizing an evidence lower bound on the data) with adversarial learning~\cite{larsen2015autoencoding,mescheder2017adversarial,srivastava2017veegan}. However, the benefits of such procedures do not come for ``free'' since we still need some form of approximate inference to get a handle on the log-likelihoods. This could be expensive, for instance combining a VAE with a GAN introduces an additional inference network that increases the overall model complexity.

Our approach sidesteps the additional complexity due to approximate inference by considering a normalizing flow model. The trade-off made by a normalizing flow model is that the generator function needs to be invertible while other generative models such as VAEs have no such requirement. On the positive side, we can tractably evaluate exact log-likelihoods assigned by the model for any data point. Normalizing flow models have been previously used in the context of maximum likelihood estimation of fully observed and latent variable  models~\cite{dinh2014nice,rezende2015variational,kingma2016improving,dinh2016density}.

The low dimensional support of the distributions learned by adversarial learning often manifests as lack of sample diversity and is referred to as mode collapse. In prior work, mode collapse is detected based on visual inspection or heuristic techniques~\cite{goodfellow2016nips,arora2017gans}. Techniques for avoiding mode collapse explicitly focus on stabilizing GAN training such as~\cite{metz2016unrolled,che2016mode,mescheder2017numerics} rather than quantitative methods based on likelihoods. 

\section{Conclusion}
As an attempt to more quantitatively evaluate generative models, we introduced Flow-GAN. It is a generative adversarial network which allows for tractable likelihood evaluation, exactly like in a flow model. Since it can be trained both adversarially (like a GAN) and in terms of MLE (like a flow model), we can quantitatively evaluate the trade-offs involved.
We observe that adversarial learning assigns very low-likelihoods to both training and validation data while generating superior quality samples. 
To put this observation in perspective, we demonstrate how a naive Gaussian mixture model can outperform adversarially learned models on both log-likelihood estimates and sample quality metrics. Quantitative evaluation methods based on AIS and KDE fail to detect this behavior and can be poor approximations of the true log-likelihood (at least for the models we considered).

Analyzing the Jacobian of the generator provides insights into the contrast between maximum likelihood estimation and adversarial learning. The latter have a tendency to learn distributions of low support, which can lead to low likelihoods. To correct for this behavior, we proposed a hybrid objective function which involves loss terms corresponding to both MLE and adversarial learning. 
The use of such models in applications requiring both density estimation and 
sample generation is an exciting direction for future work.

\section*{Acknowledgements}
We are thankful to Ben Poole and Daniel Levy for helpful discussions. This research was supported by a Microsoft Research PhD fellowship in machine learning for the first author, NSF grants $\#1651565$, $\#1522054$, $\#1733686$, a Future of Life Institute grant, and Intel.
\bibliographystyle{aaai}
\bibliography{refs}
\newpage
\section*{Appendices}
\begin{appendices}
\section{Experimental setup details}\label{app:exp_setup}
\paragraph{Datasets.} The MNIST dataset contains $50,000$ train, $10,000$ validation, and $10,000$ test images of dimensions $28\times 28$~\cite{lecun2010mnist}.  The CIFAR-10 dataset contains $50,000$ train and $10,000$ test images of dimensions $32 \times 32 \times 3$ by default~\cite{krizhevsky2009learning}. We held out a random subset of $5,000$ training set images as validation set. 

Since we are modeling densities for discrete datasets (pixels can take a finite set of values ranging from $1$ to $255$), the model can assign arbitrarily high log-likelihoods to these discrete points. Following \citeauthor{uria2013rnade}~\shortcite{uria2013rnade}, we dequantize the data by adding uniform noise between $0$ and $1$ to every pixel. Finally, we scale the pixels to lie in the range $[0,1]$. 

\paragraph{Model priors and hyperparameters.} The Flow-GAN architectures trained on MNIST and CIFAR-10 used a logistic and an isotropic prior density respectively consistent with prior work~\cite{dinh2014nice,dinh2016density}. Hyperparameter details for learning all the Flow-GAN models are included in the README of the code repository: \texttt{https://github.com/ermongroup/flow-gan}

\begin{figure}[ht]
\centering
\begin{subfigure}[b]{0.48\columnwidth}
\centering
\includegraphics[width=\columnwidth]{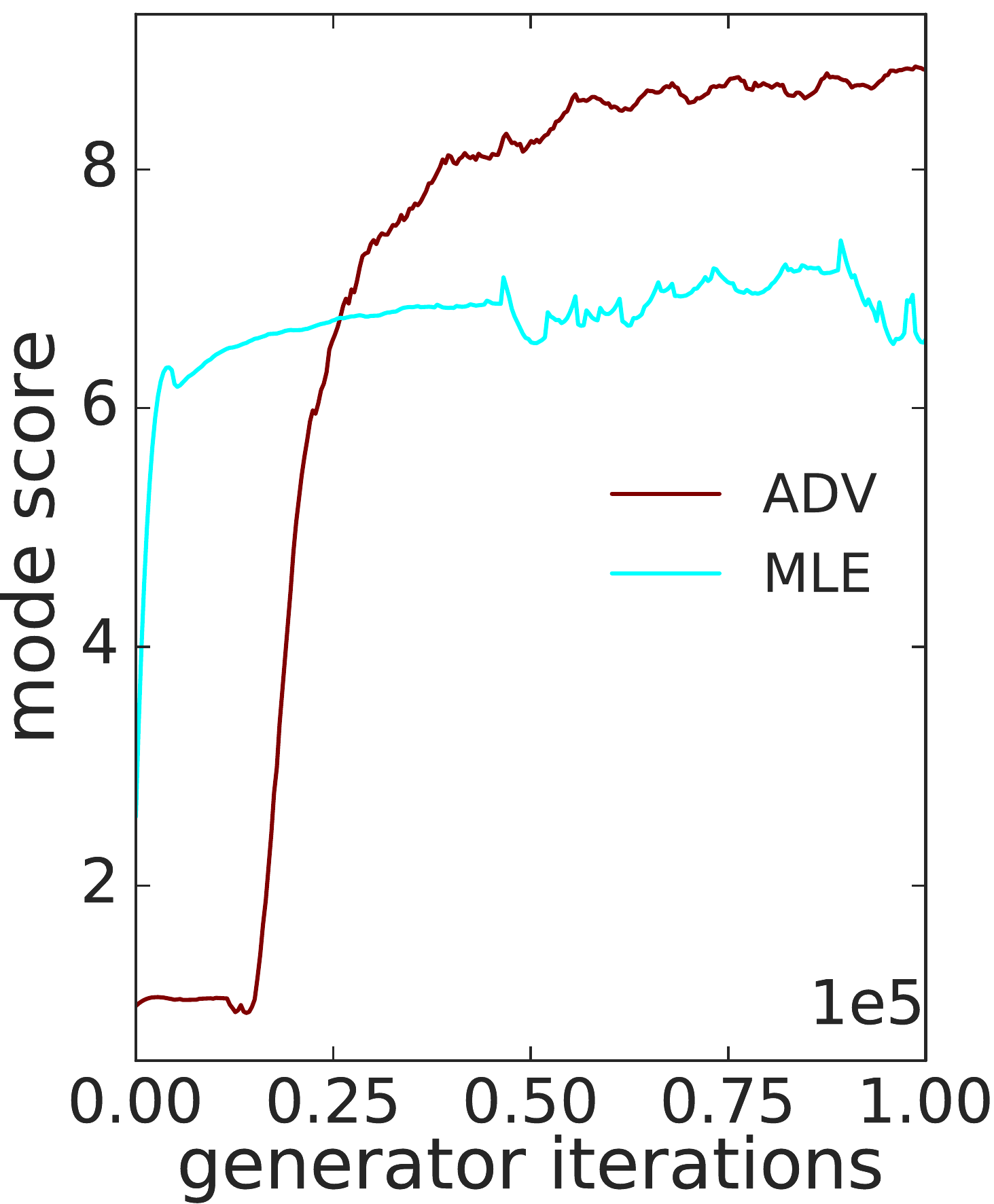}
\caption{MNIST (MODE)}
\end{subfigure}
\begin{subfigure}[b]{0.48\columnwidth}
\centering
\includegraphics[width=\columnwidth]{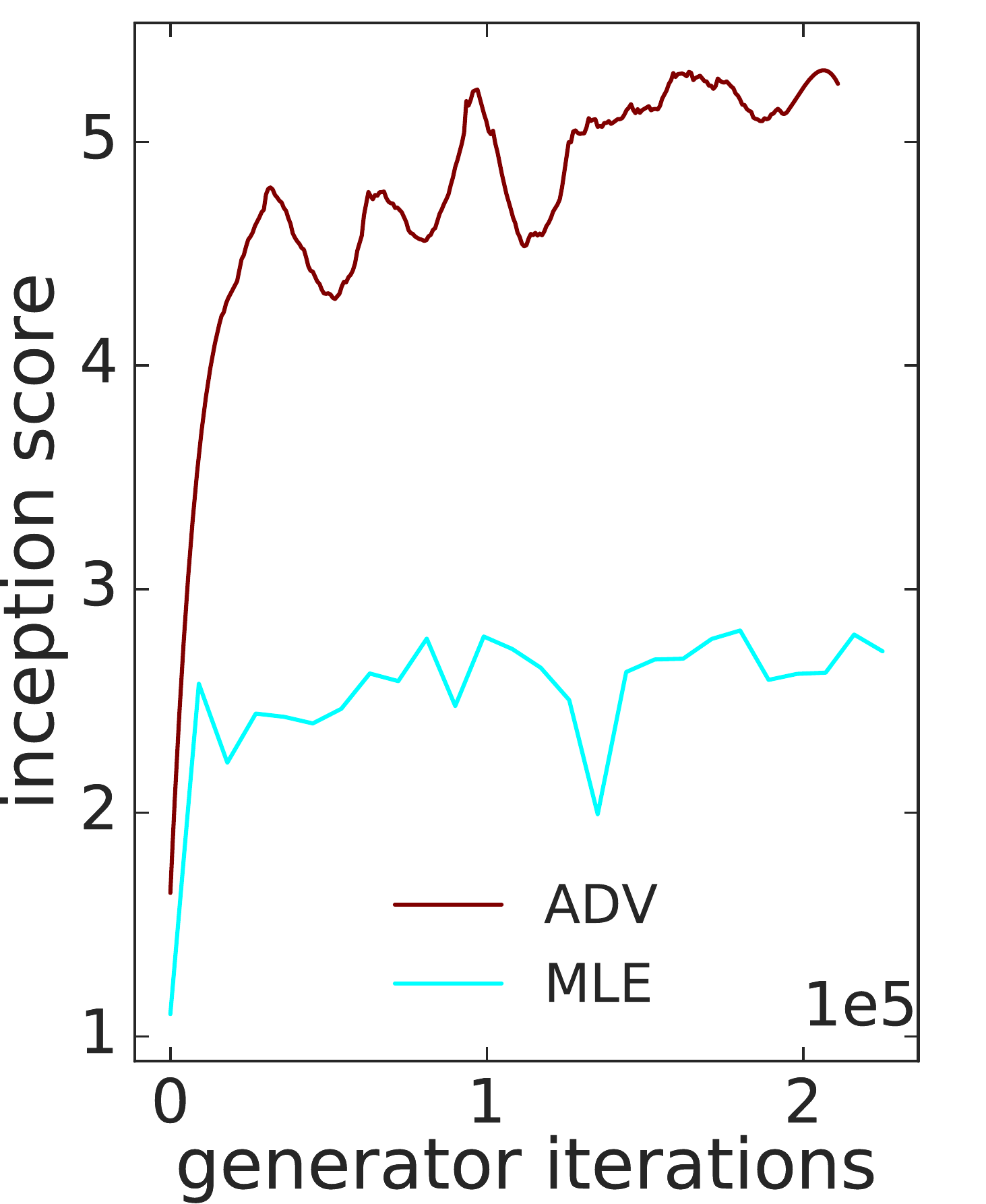}
\caption{CIFAR-10 (Inception)}
\end{subfigure}
\caption{Sample quality curves during training.}\label{fig:cifar_mnist_sample}
\end{figure}

\section{Sample quality}\label{app:sample_quality}
The progression of sample quality metrics for MLE and ADV objectives during training is shown in Figures~\ref{fig:cifar_mnist_sample} (a) and (b) for MNIST and CIFAR-10 respectively. Higher scores are reflective of better sample quality. ADV (maroon curves) significantly outperform MLE (cyan curves) with respect to the final MODE/Inception scores achieved. 
\newpage

\begin{figure}[ht]
\centering
\begin{subfigure}[b]{0.80\columnwidth}
\centering
\includegraphics[width=\columnwidth]{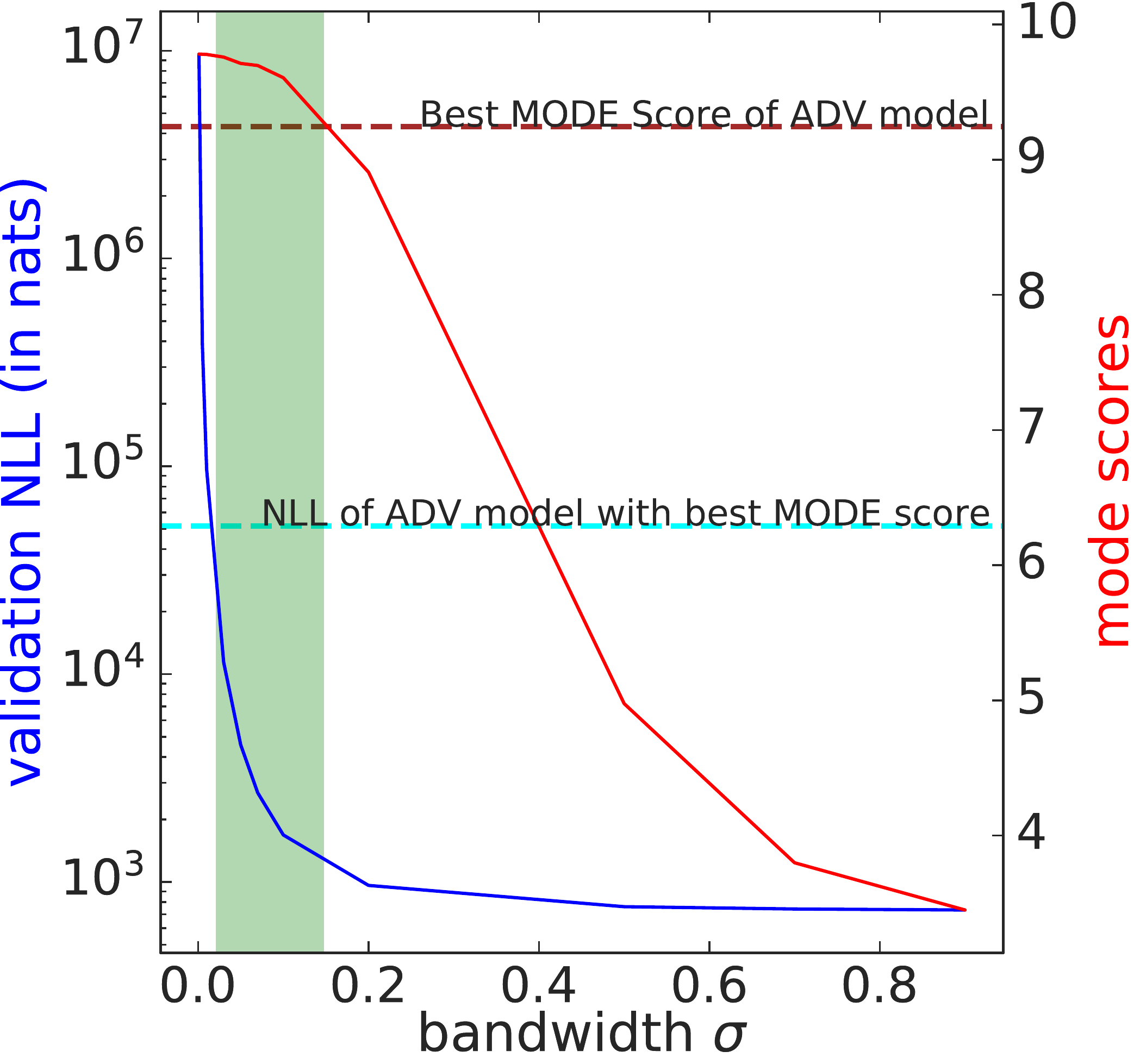}
\end{subfigure}
\caption{Gaussian Mixture Models outperform adversarially learned models on both held-out log-likelihoods and sampling metrics on MNIST (\textbf{green shaded region}).}
\label{fig:gmm_mnist}
\end{figure}

\begin{figure}[ht]
\centering
\begin{subfigure}[b]{0.33\textwidth}
\centering
\includegraphics[width=0.7\columnwidth]{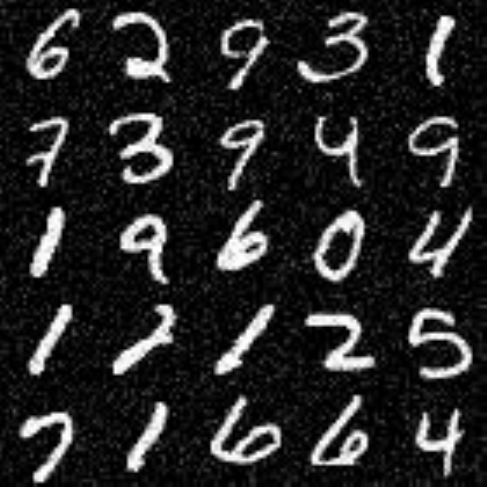}
\caption{$\sigma=0.1$}
\end{subfigure}
\begin{subfigure}[b]{0.33\textwidth}
\centering
\includegraphics[width=0.7\columnwidth]{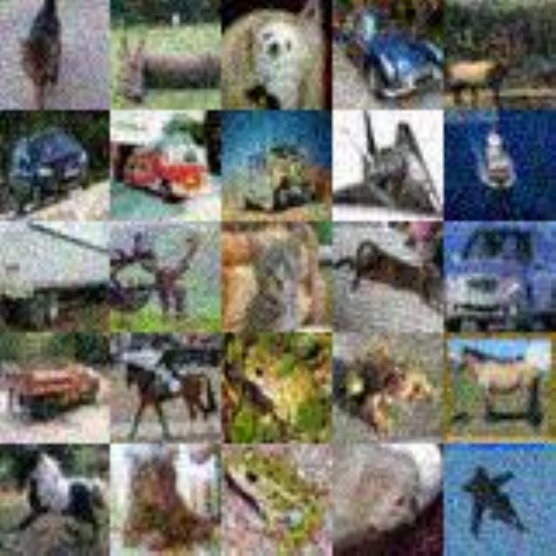}
\caption{$\sigma=0.07$}
\end{subfigure}
\caption{Samples from the Gaussian Mixture Model baseline for MNIST (\textbf{top}) and  CIFAR-10 (\textbf{bottom}) with better MODE/Inception scores than ADV models.}\label{fig:gmm_samples}
\end{figure}

\section{Gaussian mixture models}\label{app:gmm}
The comparison of GMMs with Flow-GANs trained using adversarial learning is shown in Figure~\ref{fig:gmm_mnist}.
Similar to the observations made for CIFAR-10, the simple GMM outperforms the adversarially learned model with respect to NLLs and sample quality metrics for any bandwidth parameter within the green shaded area. 

The samples obtained from the GMM are shown in Figure~\ref{fig:gmm_samples}. Since the baseline is fitting Gaussian densities around every training point, the samples obtained for relatively small bandwidths are of high quality. Yet, even the held-out likelihoods for these bandwidths are better than those of ADV models with the best MODE/Inception scores. 

\begin{figure}[ht]
\begin{subfigure}[b]{\columnwidth}
\centering
\includegraphics[width=0.75\columnwidth]{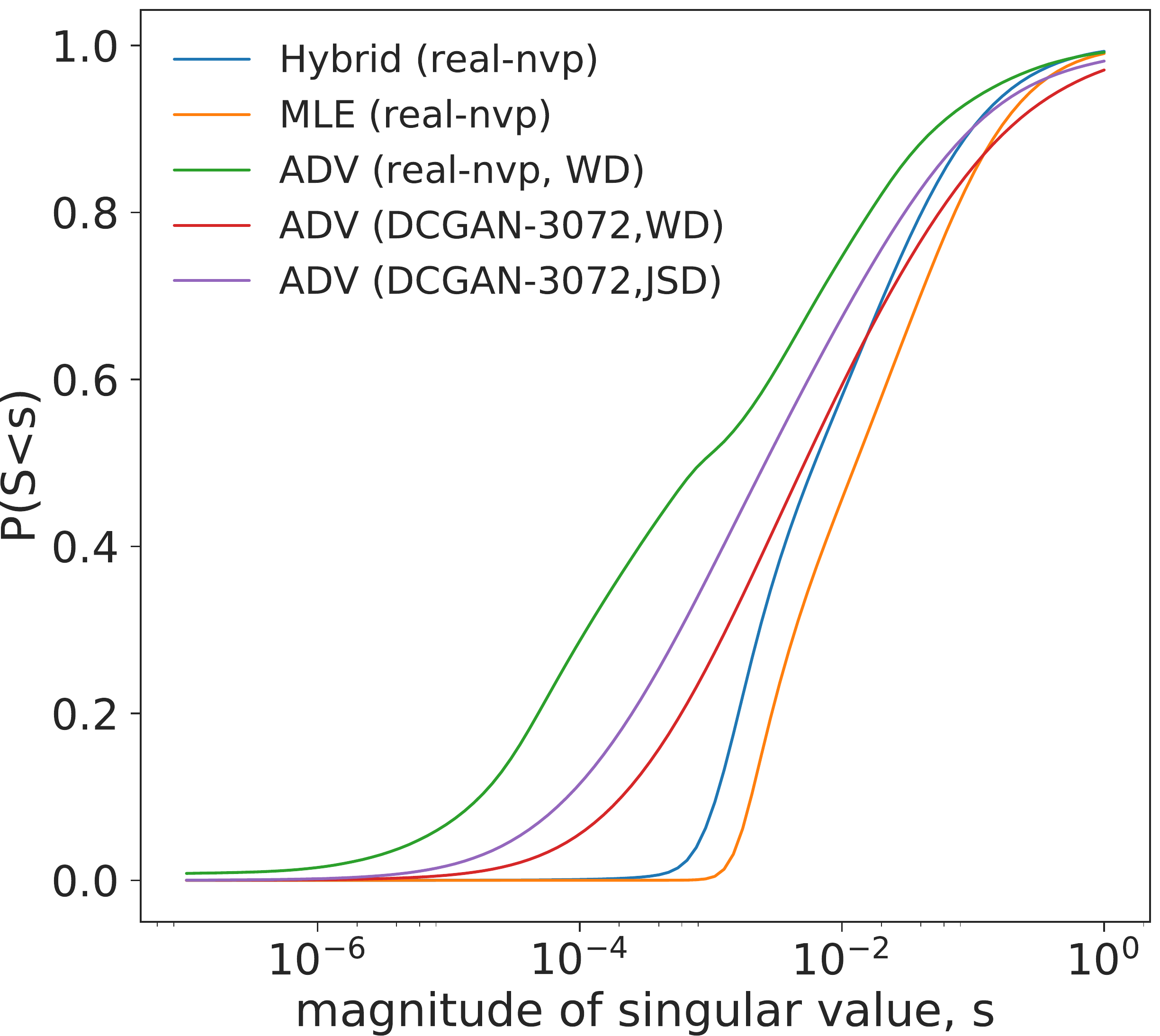}
\end{subfigure}
\caption{CDF of singular values magnitudes for the Jacobian of the generator function on CIFAR-10.}\label{fig:jac_cifar}
\end{figure}

\section{Explaining log-likelihood trends}\label{app:jac}

The CDF of singular value magnitudes for the CIFAR-10 dataset in Figure~\ref{fig:jac_cifar} again suggests that the Jacobian matrix for the generator function is ill-conditioned for the ADV models (green, red, purple curves) since the distributions have a large spread. Using a hybrid objective (blue curves) can correct for this behavior with the distribution of singular values much more concentrated similar to MLE (orange curves). 

The log determinant of the Jacobian for the MLE, ADV, and Hybrid models are $-12818.84,-21848.09,-14729.51$ respectively reflecting the trend ADV$<$Hybrid$<$MLE, providing further empirical evidence to suggest that adversarial training shows a strong preference for learning distributions with smaller support.
\end{appendices}
\end{document}